\documentclass[twocolumn, a4paper,10pt]{article}
\usepackage[top=2.5cm, bottom=2.5cm, left=2.0cm, right=2.0cm,
columnsep=0.8cm]{geometry}
\usepackage{enumitem}
\usepackage[hidelinks]{hyperref}
\usepackage{boxedminipage}
\usepackage{nopageno}
\usepackage{graphicx}%插入图片的宏包
\usepackage{float} %设置图片浮动位置的宏包
\usepackage[comma, sort]{natbib}%插入reference的宏包
\usepackage[font=it]{caption}
\usepackage[usenames,dvipsnames]{xcolor}
\usepackage{listings}
\usepackage{caption}
\usepackage{subcaption}
\usepackage{amsmath}
\usepackage{amssymb}
\usepackage{mathrsfs}
\usepackage{array}
\usepackage{booktabs}
\usepackage{color}
\usepackage{indentfirst}
\usepackage[labelfont=normalfont,textfont=normalfont]{caption}

\setlist{itemsep=-0.1cm,topsep=0.1cm,labelsep=0.3cm}
\setenumerate{leftmargin=*}
\makeatletter
\renewcommand\title[1]{\gdef\@title{\fontsize{12pt}{2pt}\bfseries{#1}}}
\makeatletter
\renewcommand\section{\@startsection{section}{1}{\z@}{3pt}{3pt}{\normalfont\large\bfseries}}
\makeatletter
\renewcommand\subsection{\@startsection{subsection}{1}{\z@}{\z@}{\z@}{\normalfont\normalsize\bfseries}}
\makeatletter
\renewcommand\subsection{\@startsection{subsection}{1}{\z@}{\z@}{0.1pt}{\normalfont\normalsize\bfseries}}

\title{%
	MLRIP: Pre-training a military language representation model with informative factual knowledge and professional knowledge base															
} 																		
\author{																% Line 3
	Hui Li$^1$ and Xuekang Yang$^1$ %Xin Zhao$^2$, Lin Yu$^1$, Jiping Zheng$^3$ and Wei Sun$ ^4 $\\ 
	\\															
	$^1$School of Automation, Nanjing University of Science and Technology, Nanjing, China\\ 																				% Line 5
	\phantom{Line 9}} 																																									% Line 10
\date{\vspace{-0.5cm}}	% remove default date and replace the Blank 10th line														
\begin{document}
	
	\maketitle
	
	\section*{Abstract}	% Section headings need to be upper and lower case.
	\addtocounter{section}{1}
	\noindent
	Incorporating structured knowledge into pre-trained language models has demonstrated significant benefits for domain-specific natural language processing tasks, particularly in specialized fields like military intelligence analysis. Existing approaches typically integrate external knowledge through masking techniques or fusion mechanisms, but often fail to fully leverage the intrinsic \textbf{tactical associations} and factual information within input sequences, while introducing uncontrolled noise from unverified external sources. To address these limitations, we present MLRIP (Military Language Representation with Integrated Prior), a novel pre-training framework that introduces a hierarchical knowledge integration pipeline combined with a dual-phase entity substitution mechanism. Our approach specifically models \textbf{operational linkages} between military entities, capturing critical dependencies such as command, support, and engagement structures. Comprehensive evaluations on military-specific NLP tasks show that MLRIP outperforms existing BERT-based models by substantial margins, establishing new state-of-the-art performance in military entity recognition, typing, and \textbf{operational linkage} extraction tasks while demonstrating superior operational efficiency in resource-constrained environments.\\	 	
	
	\section*{1 Introduction}
	\noindent
	The rapid advancement of pre-trained language representation models has fundamentally transformed the landscape of natural language processing, enabling unprecedented performance across a wide range of downstream tasks. By leveraging large-scale heterogeneous corpora and unsupervised or weakly-supervised objectives such as masked language modeling (MLM), models like ELMo \citep{peters2018deep}, OpenAI GPT \citep{radford2018improving}, and BERT \citep{devlin2018bert} have achieved state-of-the-art results in open-domain applications including named entity recognition, relation extraction, and entity typing. These breakthroughs have spurred the development of domain-specific variants tailored to specialized fields, with BioBERT \citep{symeonidou2019transfer} excelling in biomedical texts, FinBERT \citep{araci2019finbert} in financial domains, SciBERT \citep{beltagy2019scibert} in scientific literature, and PatentBERT \citep{lee2020patent} in patent documentation, each demonstrating remarkable adaptability to their respective domains.
	
	Recent trends in natural language processing have shifted toward incorporating external knowledge sources into language models, recognizing that weakly-supervised approaches \citep{xiong2019pretrained,he2019integrating,bosselut2019comet} often outperform purely unsupervised methods. This has led to the emergence of knowledge-enhanced models such as ERNIE-Baidu \citep{sun2019ernie}, which integrates phrase and entity-level masking strategies, ERNIE-Tsinghua \citep{Zhang2019Ernie}, which aligns knowledge graphs with textual content, and K-ADAPTER \citep{wang2020k}, which employs adapter-based knowledge injection techniques. These approaches have established new benchmarks on various NLP tasks, making knowledge infusion a predominant paradigm in the field.
	
	Despite these advancements, applying existing models to military text processing presents significant challenges that remain largely unaddressed. Military corpora exhibit unique linguistic characteristics that distinguish them from general domains, including specialized terminology, complex operational relationships, and varied representation styles that are not adequately captured by models trained on general-purpose corpora like Wikipedia and BookCorpus. Furthermore, pre-training tasks designed for general texts fail to account for military-specific features such as tactical associations and operational linkages, limiting their effectiveness in defense applications. The word distributions between general and military corpora differ substantially, often complicating text mining efforts in military contexts.
	
\begin{figure}[ht]
		\centering
		\includegraphics[width=0.48\textwidth]{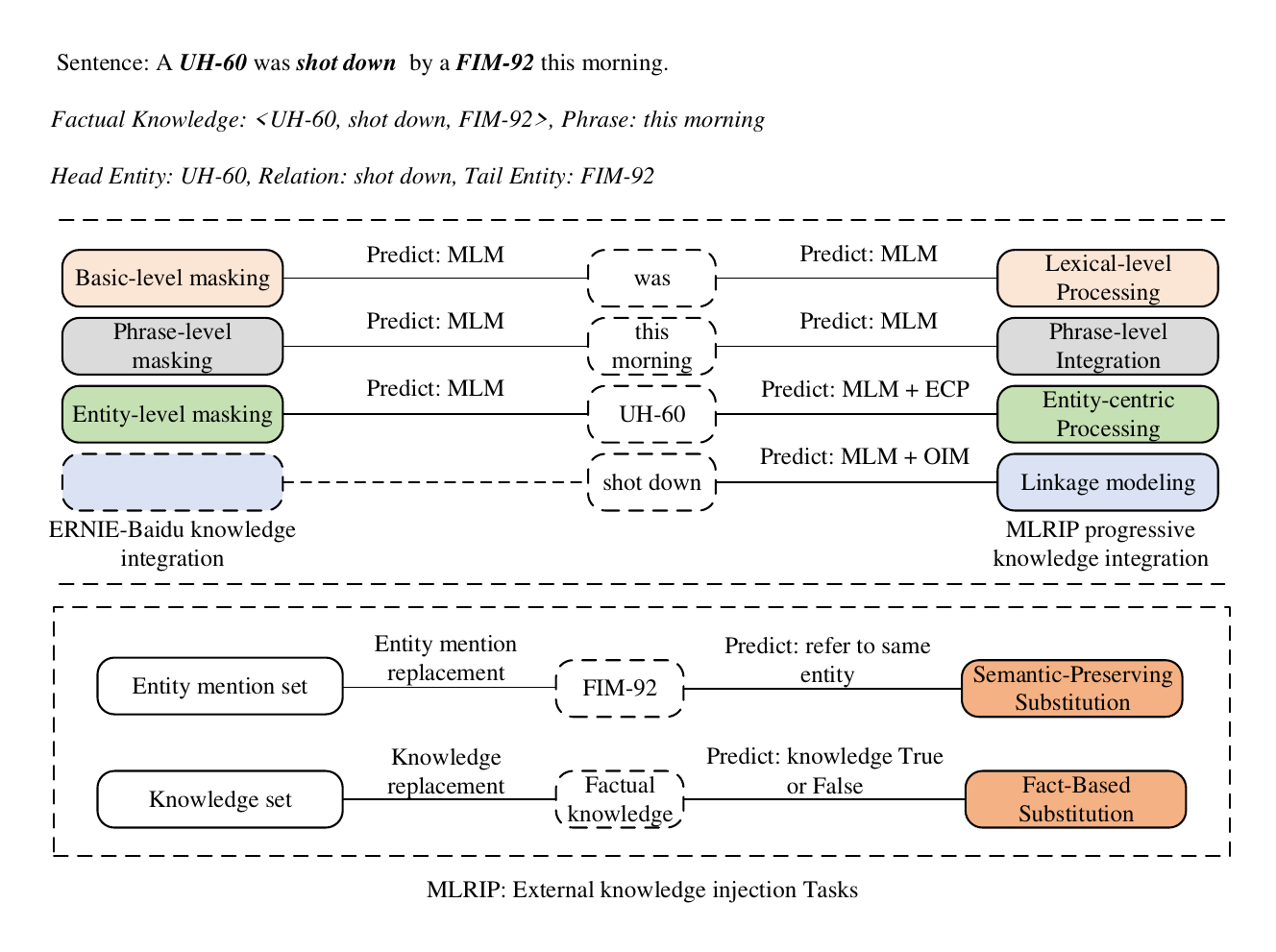}
		\caption{Comparative analysis of knowledge integration approaches between MLRIP and ERNIE-Baidu \citep{sun2019ernie}. Both frameworks employ similar methodologies for basic lexical and phrasal masking stages. However, MLRIP introduces substantial innovations in entity-centric processing through contextual prediction mechanisms that utilize available entity information and operational linkages. In contrast to ERNIE-Baidu's exclusive use of MLM for entity reconstruction, MLRIP integrates factual knowledge from the sentence context to enhance masked entity prediction. Furthermore, MLRIP extends the masking paradigm to include operational linkage-level processing, specifically designed to capture military-specific relationships and tactical associations between entities. The framework additionally incorporates external domain knowledge through two novel mechanisms: semantic-preserving entity substitution and fact-based replacement strategies optimized for military text processing.}
		\label{fig:fig02}
		\vspace{-16pt}
	\end{figure}

	Existing knowledge integration methods face additional limitations when applied to military domains. While masked language modeling approaches \citep{devlin2018bert,liu2019roberta} capture basic semantic information, they often overlook higher-level factual knowledge contained within sentences. Knowledge masking strategies \citep{sun2019ernie,joshi2020spanbert,cui2021pre} improve the incorporation of lexical, syntactic, and semantic information but fail to fully leverage the factual content present in input sequences, as illustrated in Figure \ref{fig:fig02}. Moreover, knowledge-enhanced methods \citep{liu2021oag,he2019integrating,xiong2019pretrained} frequently neglect potential issues in knowledge sources, including inaccuracies, omissions, and other inconsistencies that can introduce noise and reduce reliability. 
	
	To address these challenges, we propose MLRIP (Military Language Representation with Integrated Prior Knowledge), a novel framework specifically designed for military text mining. Our approach modifies existing knowledge masking strategies by explicitly incorporating prior factual knowledge from input sentences to enhance the prediction of masked units, while introducing a relation-level masking technique to augment the representation of operational knowledge. Furthermore, we develop a two-stage entity replacement strategy that first employs same-entity mention substitution to capture coreference variations, followed by fact-based replacement using domain-specific knowledge bases to inject authentic military knowledge. This comprehensive approach ensures robust integration of domain-specific prior knowledge while minimizing noise and maintaining semantic consistency.
	
	To evaluate the effectiveness of MLRIP, we construct comprehensive benchmark datasets for military NLP tasks and conduct extensive experiments on entity typing and relation classification. Our results demonstrate that MLRIP significantly outperforms baseline models such as BERT and ERNIE-Baidu by leveraging lexical, syntactic, and factual knowledge within sentences, combined with insights from military knowledge bases. Additional evaluations on military named entity recognition show comparable results, while ablation studies confirm that each component of our framework contributes individually to downstream task improvement.
	
	In summary, our contributions are as follows:
	
	\begin{itemize}
		
		\setlength{\itemsep}{0pt}
		
		\setlength{\parsep}{0pt}
		
		\setlength{\parskip}{0pt}
		
		\item We propose MLRIP, a novel framework for military text representation learning that introduces a progressive knowledge integration strategy combined with a dual-phase entity substitution mechanism.
		
		\item Our model significantly advances the state of the art in military NLP tasks, including entity typing and relation extraction, by effectively capturing military-specific operational linkages and tactical associations.
		
		\item We construct and release various benchmark datasets for military text mining, providing valuable resources for future research in defense-oriented natural language processing.
		
	\end{itemize}

	%---------------------------------------------------------------------------------------

	\section*{2 Related Work}
	The integration of external knowledge—whether from knowledge graphs (KGs), domain-specific datasets, additional annotations, or professional knowledge bases—has emerged as a powerful paradigm for enhancing language representation models. Such knowledge encapsulates human expertise and serves as valuable prior information to improve model performance on downstream tasks. Early work in this area focused on entity-level enhancements, with ERNIE-Baidu \citep{sun2019ernie} introducing phrase-level and entity-level masking strategies to help models capture syntactic, semantic, and dependency information across local and global contexts. Similarly, ERNIE-Tsinghua \citep{Zhang2019Ernie} incorporated knowledge graphs into BERT by aligning Wikipedia entities with Wikidata triples, enabling simultaneous learning of lexical, syntactic, and knowledge-based information.
	
	Subsequent approaches explored more structured knowledge integration. BERT-MK \citep{he2020bertmk} treated KG sub-graphs as holistic units and modeled them alongside aligned text to preserve structural information, facilitating knowledge generalization. K-ADAPTER \citep{wang2020k} proposed a plug-in architecture to inject knowledge into large pre-trained models, storing different knowledge types in separate adapters to maintain modularity. Weakly supervised methods also gained traction; WKLM \citep{xiong2019pretrained} introduced a pretraining objective that replaces entity mentions with same-type alternatives and predicts replacement events, forcing models to incorporate real-world entity knowledge. OAG-BERT \citep{liu2021oag} integrated heterogeneous knowledge from academic graphs by serializing structural knowledge into text, allowing models to learn knowledge-text alignments autonomously. ERICA \citep{qin2020erica} explicitly modeled relational facts through entity and relation discrimination tasks using contrastive learning, improving understanding of entities and relations.
	
	Recent advances have further refined knowledge integration techniques. LinkBERT \citep{yasunaga2022linkbert} leverages document links to pretrain language models, capturing inter-document relationships and enhancing contextualized representations. CoLAKE \citep{sun-etal-2020-colake} employs contextualized knowledge embeddings to jointly learn language and knowledge representations using a unified word-knowledge graph, improving performance on knowledge-driven tasks. Additionally, K-PLM \citep{wang-etal-2022-knowledge} introduces a knowledge-prompting paradigm with self-supervised tasks to enhance natural language understanding in both full-resource and low-resource settings. These methods demonstrate the ongoing evolution toward more dynamic, context-aware, and modular knowledge integration frameworks.
	
	%---------------------------------------------------------------------------------------

	\section*{3 Methodological Framework}
	
	\subsection*{3.1 Formalization and Notation}
	We establish a formal framework for military text processing by defining a sentence as an ordered sequence of linguistic units \(\mathcal{S} = \{u_1, u_2, \dots, u_N\}\), where \(N\) represents the sequence length and each \(u_i\) denotes a fundamental textual element. For Chinese language processing, we utilize character-level tokens, while for English we employ word-level tokens. The encoded representation is formulated as \(\mathcal{E}(\mathcal{S}) = \{\mathbf{z}_1, \mathbf{z}_2, \dots, \mathbf{z}_N\}\), where each \(\mathbf{z}_i\) corresponds to the contextualized embedding of the respective linguistic unit. The comprehensive token inventory is designated as \(\mathcal{V}\), and the military entity lexicon containing all known entities is denoted as \(\mathcal{M}\), encompassing 148 distinct entity categories and 5,775 unique entity instances with structured metadata stored in JSON format.
	
	\subsection*{3.2 Architectural Foundation}
	Our methodological approach is grounded in the transformer architecture \citep{vaswani2017attention}, implementing a deep bidirectional encoder configuration that has demonstrated exceptional performance across diverse language representation tasks \citep{devlin2018bert,liu2019roberta}. Specifically, we deploy a 12-layer architecture with 12 parallel attention mechanisms and a latent dimension of 768, yielding approximately 110 million trainable parameters—architecturally aligned with BERT\(_{\text{base}}\) \citep{devlin2018bert}—ensuring seamless integration with existing downstream applications while incorporating specialized military domain adaptations.
	
\begin{figure}[ht]
		\centering
		\includegraphics[width=0.48\textwidth]{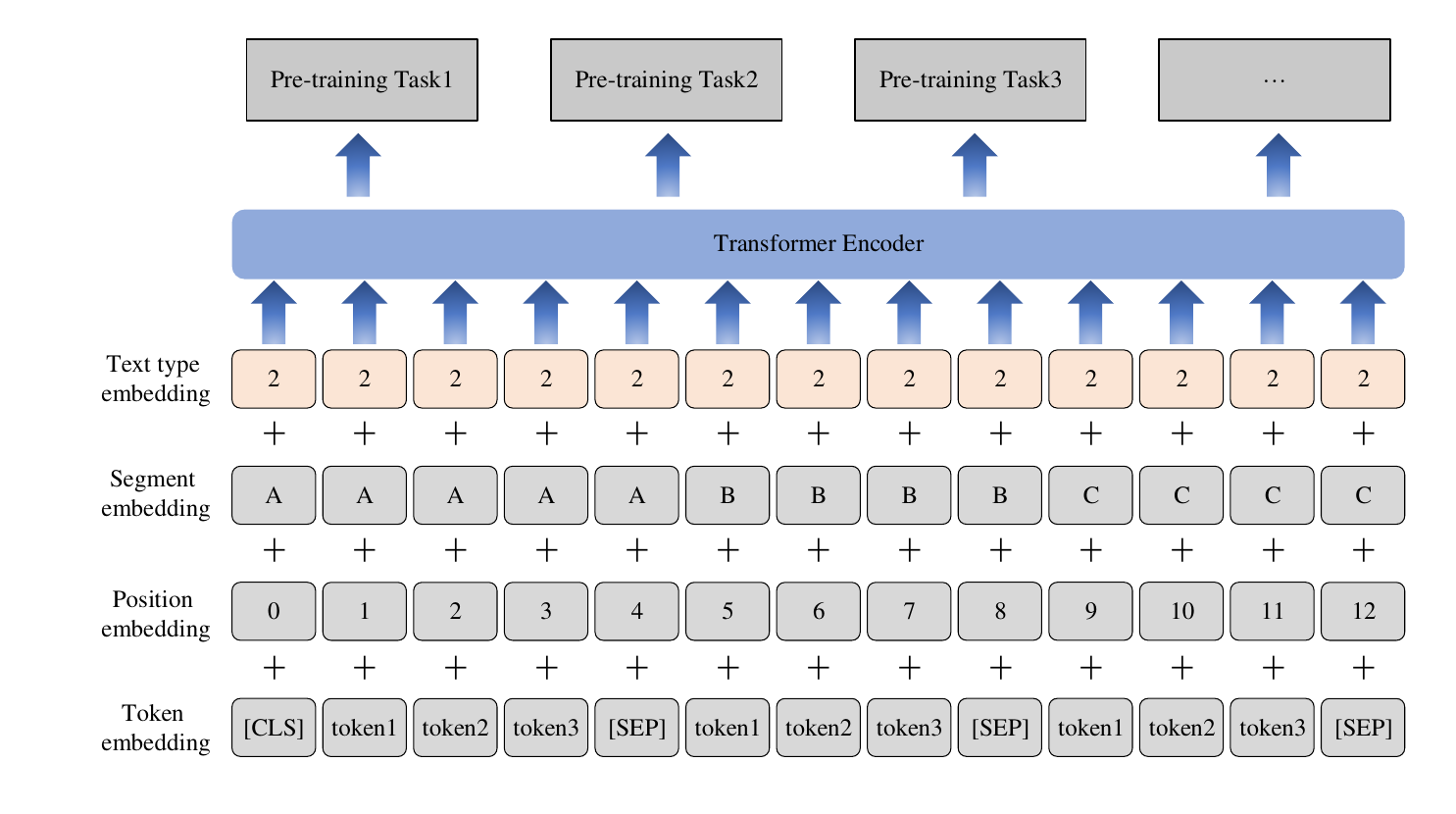}
		\caption{Comprehensive architectural overview of the MLRIP framework, illustrating the integrated components for military text representation learning, including the enhanced embedding mechanism and progressive knowledge assimilation stages.}
		\label{fig:fig01}
		\vspace{-16pt}
	\end{figure}

	The model processes input sequences through successive transformer modules that utilize self-attention mechanisms to capture intricate linguistic patterns, including lexical, syntactic, and semantic relationships particularly relevant to military contexts. Each module comprises multi-head self-attention layers followed by position-wise feedforward networks, with residual connections and layer normalization consistently applied \citep{vaswani2017attention}. The comprehensive architectural framework, which integrates these components with enhanced embedding schemes and progressive knowledge integration stages, is visualized in Figure~\ref{fig:fig01}.
	
	\subsection*{3.3 Enhanced Embedding Mechanism}
	We introduce an advanced embedding strategy that incorporates multiple information sources to better capture military-specific characteristics. Beyond conventional token, segment, and position embeddings employed in standard models \citep{devlin2018bert}, we integrate a novel document-type embedding that identifies specific categories of military documents (e.g., operational reports, intelligence assessments, technical specifications). This enables the model to dynamically adjust its representations based on document characteristics and military operational context.
	
	Formally, the composite embedding representation for each linguistic unit is computed as:
	\[
	\mathbf{e}_i = \mathbf{e}^{(t)}_i + \mathbf{e}^{(s)}_i + \mathbf{e}^{(p)}_i + \mathbf{e}^{(d)}_i
	\]
	where each component is learned during pre-training, with superscripts denoting: (t) token-level, (s) segment-level, (p) position-aware, and (d) document-type embeddings. The document-type embedding proves particularly valuable for military applications where document provenance and classification significantly influence interpretation and analytical outcomes.
	
	\subsection*{3.4 Progressive Knowledge Integration Framework}
	We develop a multi-stage knowledge integration methodology that systematically incorporates military domain knowledge through four progressive learning phases. This structured approach enables the model to capture increasingly sophisticated linguistic and domain-specific patterns.
	
	\subsubsection*{3.4.1 Lexical-Level Processing}
	In the foundational stage, we concentrate on fundamental lexical representations using masked token prediction analogous to standard language models \citep{devlin2018bert}. However, we enhance this process by incorporating military-specific lexical preferences and terminological distributions. Specifically, 15\% of tokens are randomly masked, with the model trained to reconstruct original tokens based on contextual information. This foundation enables the model to acquire essential lexical patterns and military terminology.
	
	\subsubsection*{3.4.2 Phrasal-Level Integration}
	Building upon lexical foundations, we extend the masking strategy to multi-word expressions prevalent in military contexts, such as equipment designations ("air-to-surface missile") and operational terminology ("close combat support"). Similar to ERNIE-Baidu \citep{sun2019ernie}, we treat these phrases as unified units during masking, enabling the model to learn cohesive semantic representations of military phrases. This approach captures specialized syntactic structures frequently encountered in military documentation.
	
	\subsubsection*{3.4.3 Entity-Centric Processing}
	Military operations frequently involve specific entities (personnel, equipment, locations, organizations) requiring precise identification and representation. In this stage, we mask complete entity mentions irrespective of token length, compelling the model to develop robust entity representations. For instance, given the sentence "The F-35 Lightning II conducted surveillance operations," we would mask the entire aircraft designation "F-35 Lightning II" as a singular unit.
	
	The entity masking process incorporates factual knowledge from the context to enhance prediction accuracy. For a factual triple \(\langle h, r, t \rangle\) where head entity \(h\) is masked, we compute representations for the relation \(r\) and tail entity \(t\) through mean pooling of their constituent tokens:
	\[
	\mathbf{v}_t = \frac{1}{L_t} \sum_{j=\text{start}(t)}^{\text{end}(t)} \mathbf{z}_j
	\]
	\[
	\mathbf{v}_r = \frac{1}{L_r} \sum_{k=\text{start}(r)}^{\text{end}(r)} \mathbf{z}_k
	\]
	where \(L_t\) and \(L_r\) denote the length of entity \(t\) and relation \(r\) respectively, and \(\mathbf{z}_j\) represents the contextual embedding of the \(j\)-th token from the encoder output.
	
	We subsequently compute the representation for each masked token in \(h\) using a transformation function:
	\[
	\mathbf{h}_i = g(\mathbf{v}_t, \mathbf{v}_r, \mathbf{p}_{i+\text{start}(h)})
	\]
	where \(\mathbf{p}\) denotes position embeddings and \(g\) is implemented as a two-layer feedforward network with GeLU activations and layer normalization, following architectural principles from SpanBERT \citep{joshi2020spanbert}.
	
	\begin{figure}[ht]
		\centering
		\includegraphics[width=0.48\textwidth]{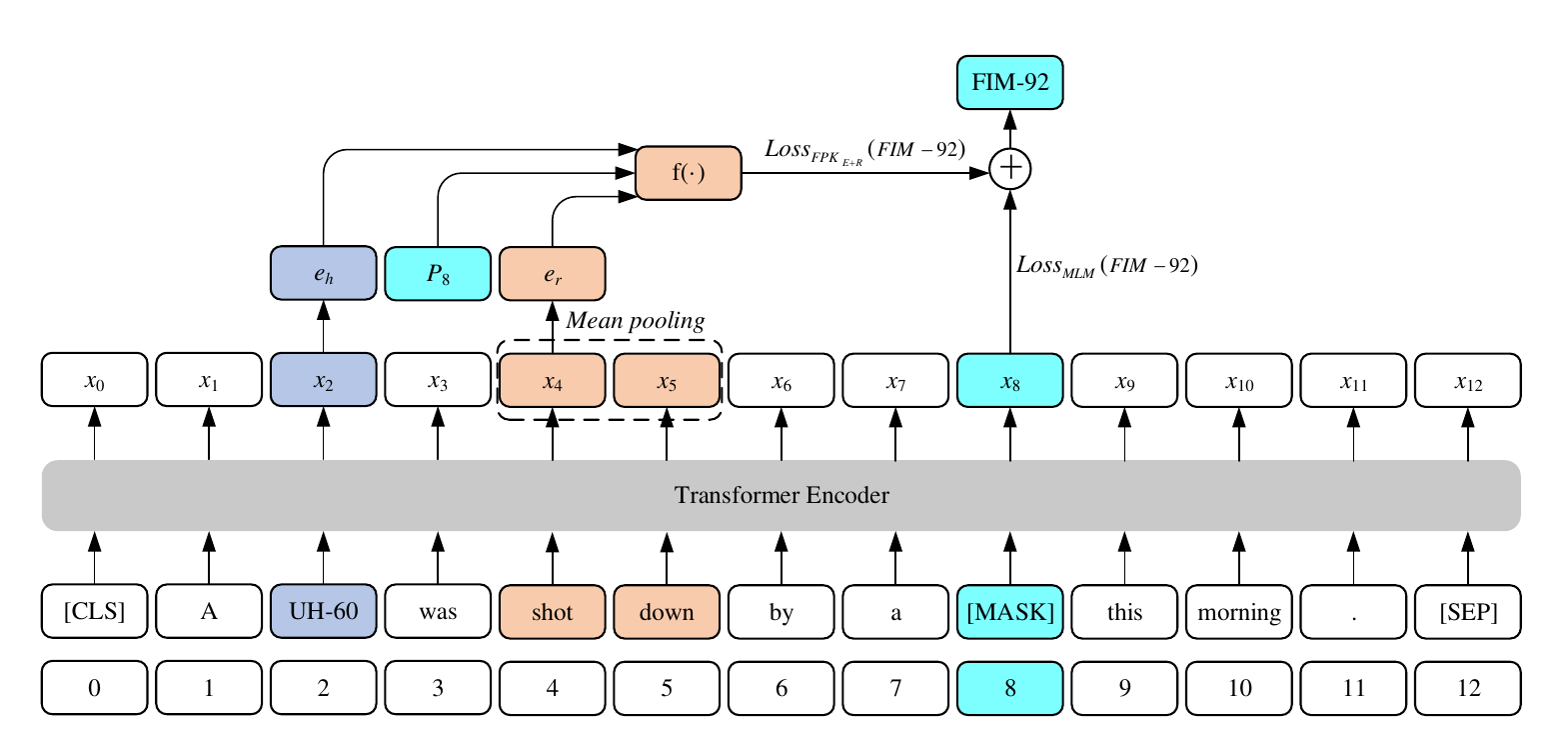}
		\caption{Entity-centric processing methodology, demonstrating how military entities are identified and processed as complete units with contextual factual knowledge integration.}
		\label{fig:fig03}
		\vspace{-16pt}
	\end{figure}
	
	\subsubsection*{3.4.4 Operational Interrelationship Modeling}
	The culminating stage focuses on capturing military-specific relationships between entities, termed "operational interrelationships." These include command hierarchies, support mechanisms, engagement protocols, and other tactical associations unique to military contexts. We mask these relational expressions and train the model to reconstruct them based on contextual entity information.
	
	To formalize this process, let \(s = \{w_1, w_2, \dots, w_n\}\) be a token sequence containing a factual triple \(\langle h, r, t \rangle\), where \(h\) denotes the head entity, \(r\) the relation, and \(t\) the tail entity. When the relation \(r\) is masked, we first encode the sequence using the transformer encoder to obtain token representations \(\text{enc}(s) = \{\mathbf{z}_1, \mathbf{z}_2, \dots, \mathbf{z}_n\}\). We then compute representations for the head and tail entities through weighted mean pooling with layer normalization:
	\[
	\mathbf{v}_h = \text{LayerNorm}\left( \frac{1}{L_h} \sum_{i=\text{start}(h)}^{\text{end}(h)} \alpha_i \cdot \mathbf{z}_i \right)
	\]
	\[
	\mathbf{v}_t = \text{LayerNorm}\left( \frac{1}{L_t} \sum_{j=\text{start}(t)}^{\text{end}(t)} \beta_j \cdot \mathbf{z}_j \right)
	\]
	where \(L_h\) and \(L_t\) denote the lengths of entities \(h\) and \(t\), respectively, and \(\alpha_i\), \(\beta_j\) are learnable weight parameters that emphasize important tokens within entities. The relation representation \(\mathbf{v}_r\) is derived through a nonlinear transformation function that combines entity representations:
	\[
	\mathbf{v}_r = \sigma\left( \mathbf{W}_1 [\mathbf{v}_h; \mathbf{v}_t] + \mathbf{b}_1 \right)
	\]
	Here, \(\sigma\) is the GeLU activation function, \(\mathbf{W}_1\) and \(\mathbf{b}_1\) are learnable parameters, and \([\cdot; \cdot]\) denotes concatenation. For each masked token in relation \(r\), we compute its contextualized representation using a position-aware function:
	\[
	\mathbf{v}_r(k) = \text{GeLU}\left( \mathbf{W}_2 \mathbf{v}_r + \mathbf{W}_3 \mathbf{p}_{k + \text{start}(r)} + \mathbf{b}_2 \right)
	\]
	where \(\mathbf{p}\) denotes position embeddings, and \(k\) is the relative position within the relation. The model then predicts the original relation token via a softmax classifier over the vocabulary, with the loss function defined as the cross-entropy:
	\[
	\mathcal{L}_{\text{rel}} = -\sum_{k=1}^{L_r} \log P\left( r(k) \mid \mathbf{v}_r(k) \right)
	\]
	where \(L_r\) is the length of relation \(r\). This approach enables the model to learn complex military operational patterns essential for tactical situation understanding.
	
	\begin{figure}[ht]
		\centering
		\includegraphics[width=0.48\textwidth]{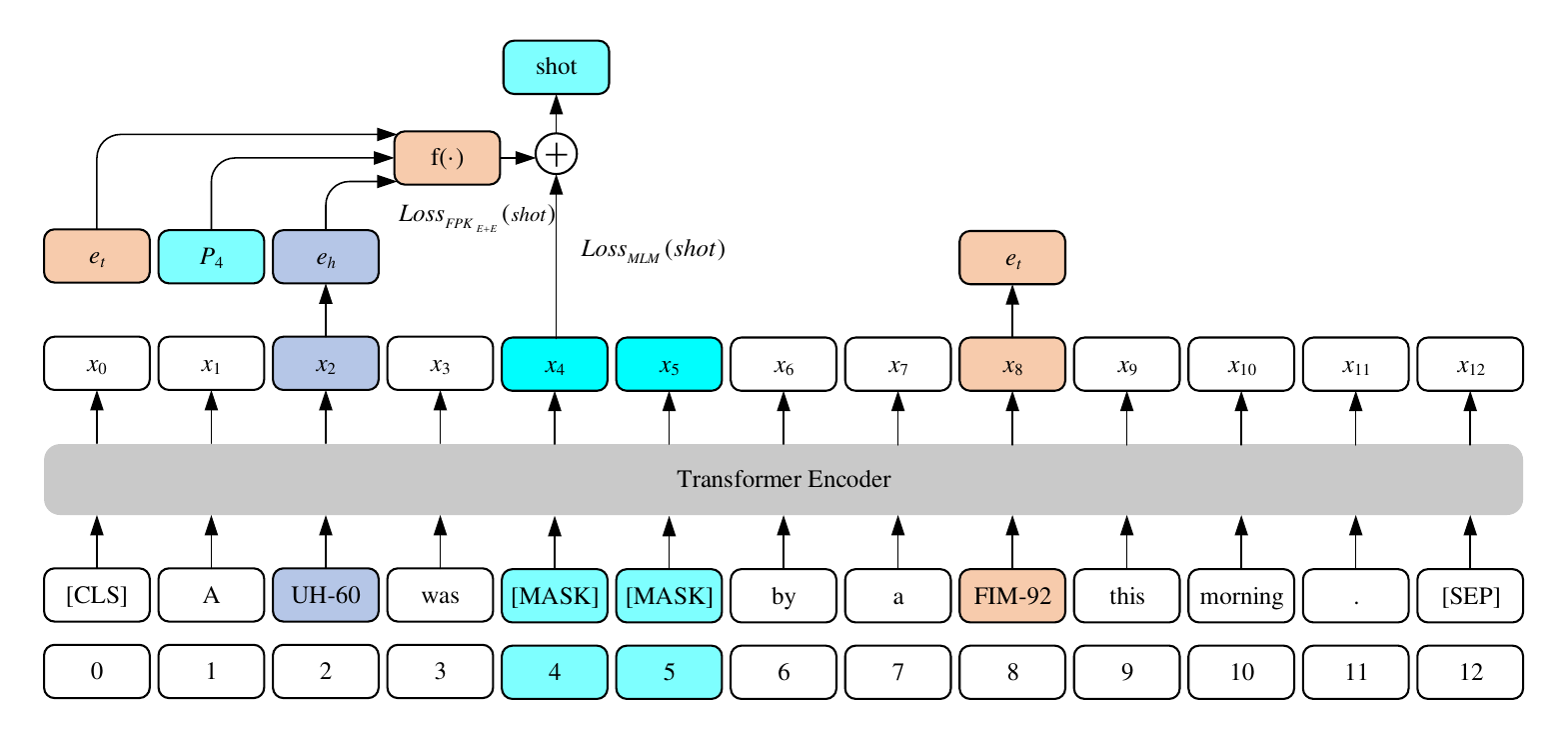}
		\caption{Operational interrelationship modeling methodology, illustrating how military-specific relationships between entities are identified and processed.}
		\label{fig:fig04}
		\vspace{-16pt}
	\end{figure}
	
	\subsection*{3.5 Knowledge Incorporation via Entity Substitution}
	To further enhance the model's military domain knowledge, we implement a dual-phase entity substitution strategy that incorporates external military knowledge while maintaining semantic consistency.
	
	\subsubsection*{3.5.1 Semantic-Preserving Substitution}
	Military entities frequently exhibit multiple valid designations (e.g., "F-35," "Lightning II," "Joint Strike Fighter"). In this phase, we replace entity mentions with semantically equivalent alternatives from our military entity lexicon \(\mathcal{M}\). The substitution probability is derived from semantic similarity computed using pre-trained word vectors:
	\[
	d_i = \| \mathbf{m}_i - \mathbf{m}_0 \|_2^2
	\]
	\[
	P(i) = \frac{\exp(-d_i)}{\sum_j \exp(-d_j)}
	\]
	where \(\mathbf{m}_i\) represents the embedding of candidate mention \(i\) and \(\mathbf{m}_0\) represents the original mention embedding. This approach enhances the model's capability to recognize entity coreference and variant designations prevalent in military texts.
	
	\subsubsection*{3.5.2 Fact-Based Substitution}
	Extending beyond semantic substitution, we incorporate factual military knowledge by replacing entities based on operational relationships documented in military knowledge repositories. For instance, we might substitute "UH-60" with "CH-47" in contexts involving heavy lift operations, based on established operational capabilities.
	
	This process employs a knowledge parser to extract factual information from text, a reasoning engine to generate militarily valid substitutions, and a verification module to ensure factual accuracy. The proportion of positive (militarily valid) to negative (invalid) substitutions is maintained at 1:2 to prevent overfitting and ensure robust learning.
	
	\begin{figure}[ht]
		\centering
		\includegraphics[width=0.48\textwidth]{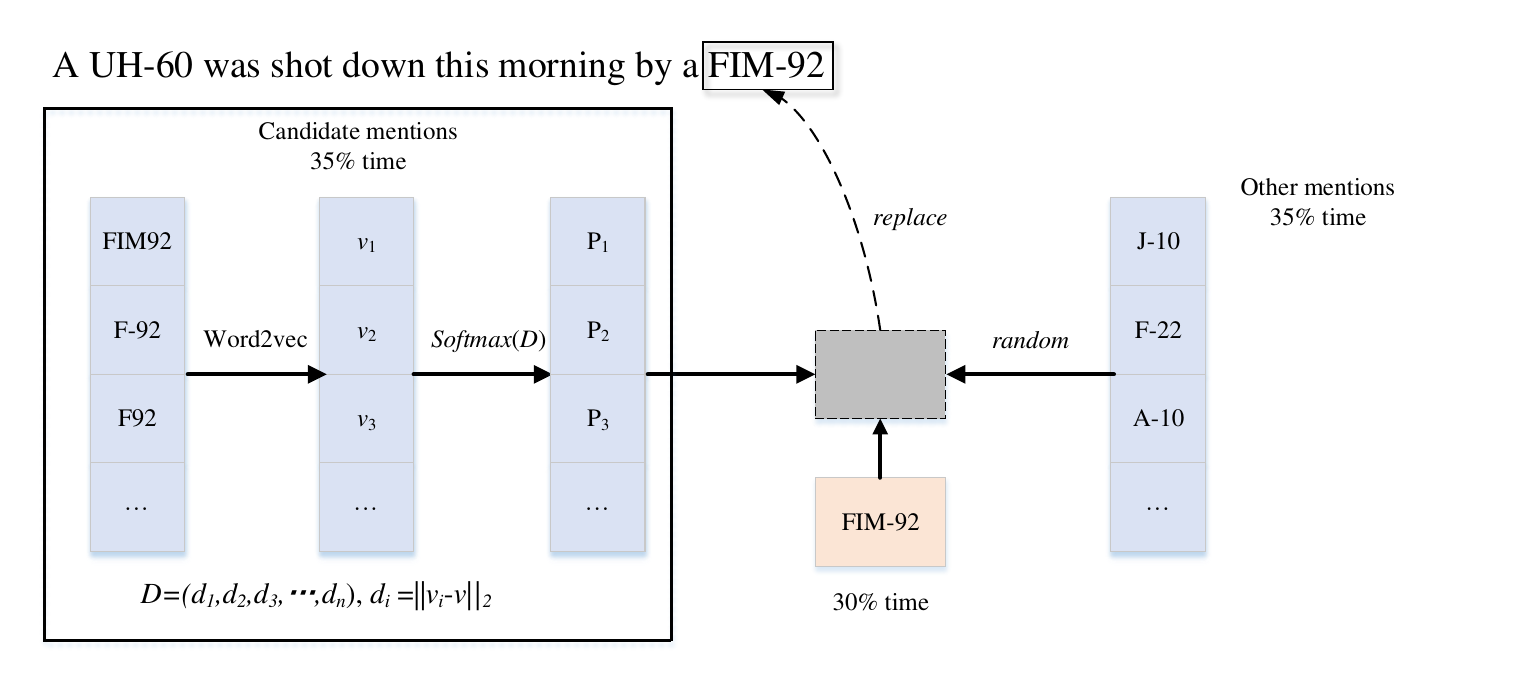}
		\caption{Entity substitution methodology, illustrating both semantic-preserving and fact-based replacement strategies for military entities.}
		\label{fig:fig05}
		\vspace{-16pt}
	\end{figure}
	
	\subsubsection*{3.5.1 Semantic-Preserving Substitution}
	Military entities frequently exhibit multiple valid designations (e.g., "F-35," "Lightning II," "Joint Strike Fighter"). In this phase, we replace entity mentions with semantically equivalent alternatives from our military entity lexicon \(\mathcal{M}\). The substitution probability is derived from semantic similarity computed using pre-trained word vectors. Specifically, we obtain entity embeddings through a pre-trained Word2Vec model trained on military corpora, with each embedding normalized to unit length to ensure scale invariance:
	
	\[
	\mathbf{m}_i = \frac{\mathbf{w}_i}{\|\mathbf{w}_i\|_2}, \quad \mathbf{m}_0 = \frac{\mathbf{w}_0}{\|\mathbf{w}_0\|_2}
	\]
	where \(\mathbf{w}_i\) and \(\mathbf{w}_0\) are raw word vectors for candidate mention \(i\) and original mention, respectively. The semantic distance is computed using squared Euclidean distance:
	\[
	d_i = \| \mathbf{m}_i - \mathbf{m}_0 \|_2^2
	\]
	The substitution probability is then derived through a softmax normalization over all candidate mentions in \(\mathcal{M}\):
	\[
	P(i) = \frac{\exp(-\gamma d_i)}{\sum_{j \in \mathcal{M}} \exp(-\gamma d_j)}
	\]
	where \(\gamma\) is a temperature parameter (set to 1.0 empirically) that controls the distribution sharpness. This approach enhances the model's capability to recognize entity coreference and variant designations prevalent in military texts by emphasizing semantically closer alternatives.
	
	\subsubsection*{3.5.2 Fact-Based Substitution}
	Extending beyond semantic substitution, we incorporate factual military knowledge by replacing entities based on operational relationships documented in military knowledge repositories. For instance, we might substitute "UH-60" with "CH-47" in contexts involving heavy lift operations, based on established operational capabilities. This process employs a knowledge parser to extract factual information from text, a reasoning engine to generate militarily valid substitutions, and a verification module to ensure factual accuracy.
	
	Formally, let \(\mathcal{K}\) denote the military knowledge base containing factual triples \(\langle e_1, r, e_2 \rangle\), where \(e_1\) and \(e_2\) are entities and \(r\) is a relation. For a given entity \(e_0\) in context, we query \(\mathcal{K}\) to retrieve substitutable entities based on relational constraints. The candidate set is defined as:
	\[
	\mathcal{S}_{\text{candidate}} = \left\{ e \middle| 
	\begin{split}
		&\langle e_0, r, e \rangle \in \mathcal{K} \quad \text{or} \\
		&\langle e, r, e_0 \rangle \in \mathcal{K} \quad \text{for some relation } r
	\end{split}
	\right\}
	\]
	The substitution probability for candidate entity \(e\) is weighted by relational strength and contextual relevance:
	\[
	P(e) = \frac{\exp(\phi(e, e_0, c))}{\sum_{e' \in \mathcal{S}_{\text{candidate}}} \exp(\phi(e', e_0, c))}
	\]
	where \(\phi(e, e_0, c)\) is a scoring function that combines knowledge base confidence and contextual similarity:
	\[
	\phi(e, e_0, c) = \lambda \cdot \text{sim}(\mathbf{v}_e, \mathbf{v}_{e_0}) + (1 - \lambda) \cdot \text{conf}(e, e_0)
	\]
	Here, \(\mathbf{v}_e\) and \(\mathbf{v}_{e_0}\) are contextualized entity embeddings from the model, \(\text{sim}(\cdot)\) is cosine similarity, \(\text{conf}(e, e_0)\) is the confidence score from the knowledge base for the relationship between \(e\) and \(e_0\), and \(\lambda\) (set to 0.7) balances contextual and knowledge factors. The proportion of positive (militarily valid) to negative (invalid) substitutions is maintained at 1:2 to prevent overfitting and ensure robust learning.
	
	\begin{figure}[ht]
		\centering
		\includegraphics[width=0.48\textwidth]{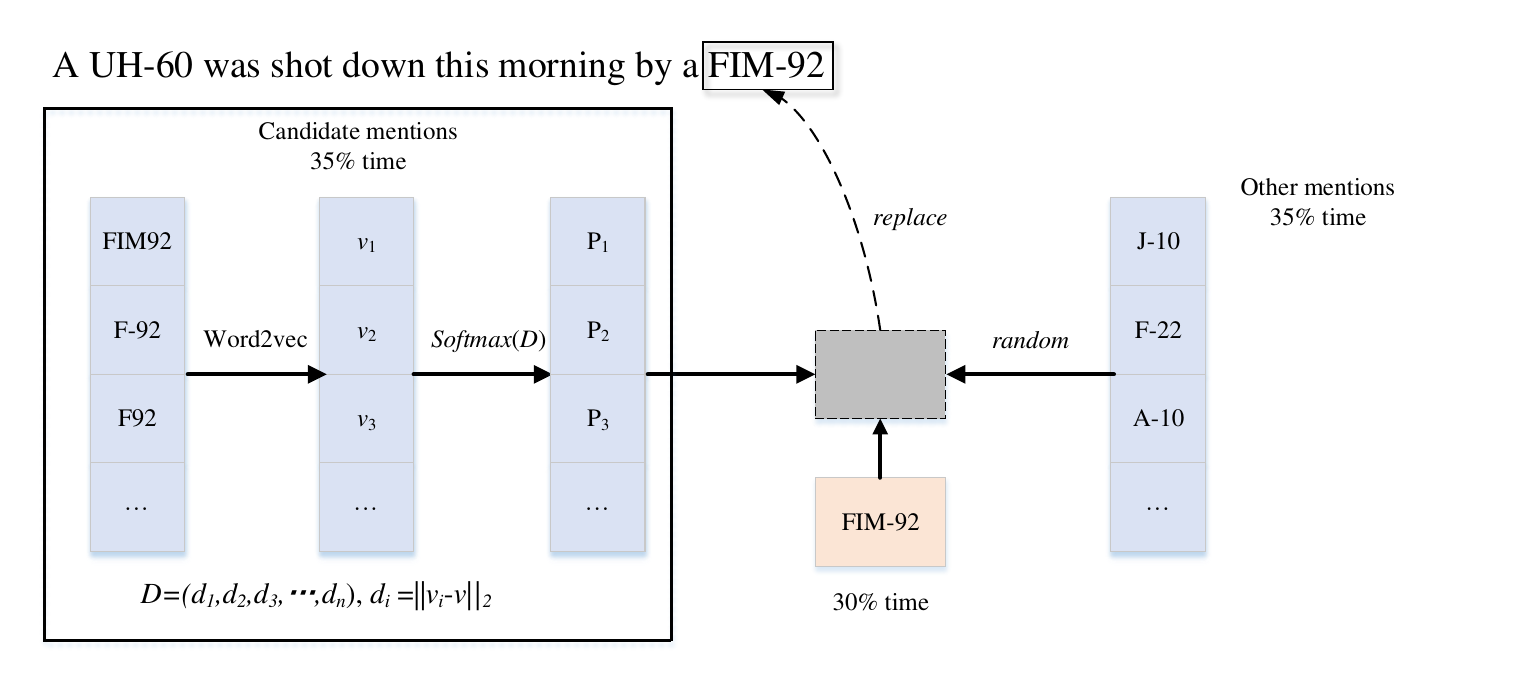}
		\caption{Entity substitution methodology, illustrating both semantic-preserving and fact-based replacement strategies for military entities.}
		\label{fig:fig05}
		\vspace{-16pt}
	\end{figure}
	
	\subsection*{3.6 Implementation and Training Specifications}
	We implement our approach using a continuous multi-task learning framework inspired by ERNIE 2.0 \citep{sun2020ernie2}. Training is conducted on 8 NVIDIA Tesla V100 GPUs with 32GB memory each. We employ a maximum sequence length of 256 tokens and a batch size of 512 to optimize training efficiency. The model is optimized using AdamW \citep{kingma2014adam} with a learning rate of 3e-5, L2 weight decay of 1e-3, and linear learning rate warmup for the initial 20\% of training steps followed by linear decay.
	
	Pre-training utilizes a diverse military corpus comprising operational documents, intelligence reports, technical manuals, and historical accounts totaling approximately 2.7 billion tokens. The composite training objective incorporates multiple loss components:
	\[
	\mathcal{L}_{\text{total}} = \alpha \mathcal{L}_{\text{mask}} + \beta \mathcal{L}_{\text{ent}} + \gamma \mathcal{L}_{\text{link}} + \delta \mathcal{L}_{\text{sub}}
	\]
	where each component corresponds to one of our knowledge integration stages, and \(\alpha, \beta, \gamma, \delta\) represent weighting coefficients. This multi-objective optimization ensures balanced learning across different dimensions of military language understanding.
	
	Complete training requires approximately 2 million steps, consuming roughly three weeks of computation time. The resultant model maintains full compatibility with BERT-based architectures while incorporating specialized military domain knowledge through our progressive integration methodology.

	\section*{4 Experimental Evaluation}
	
	\subsection*{4.1 Military Data Composition and Characteristics}
	The experimental foundation of our study relies on a meticulously curated collection of military-domain textual resources. Our training corpus encompasses five distinct categories of professional military documentation: operational directives, intelligence assessments, tactical scenario descriptions, technical specifications, and simulation exercise records. Complementing these, we incorporated seven types of openly available military content from digital sources: defense news portals, tactical gaming communities, equipment evaluation forums, professional military blogs, strategic analysis articles, policy discussion platforms, and Chinese military reference materials.
	
	Previous investigations in domain adaptation \citep{li2022fusion} have demonstrated that military texts exhibit unique linguistic characteristics that distinguish them from general corpora. Formal doctrinal publications employ standardized terminology and structured presentation formats, while tactical communications and intelligence reports utilize abbreviated forms, code designations, and context-dependent expressions. These variations necessitate specialized processing approaches that conventional models often fail to address adequately.
	
	\begin{table*}[ht]
		\centering
		\small
		\caption{Statistical overview of military text evaluation benchmarks}
		\label{tab:benchmark_stats}
		\begin{tabular}{@{}lcccccc@{}}
			\toprule
			\textbf{Dataset} & \textbf{Training} & \textbf{Development} & \textbf{Test} & \textbf{Categories} & \textbf{Entity Count} \\
			\midrule
			FGMET & 256,000 & 32,000 & 32,000 & 71 & -- \\
			MNER & 24,960 & 3,120 & 3,120 & 12 & 16,729 \\
			LCRECM & 61,458 & 21,098 & 15,280 & 132 & -- \\
			\bottomrule
		\end{tabular}
		\vspace{-4pt}
	\end{table*}
	
	Given the absence of standardized evaluation frameworks for military NLP tasks, we developed three comprehensive benchmark datasets following established practices in general-domain NLP evaluation \citep{wang2018glue, rajpurkar2016squad}. The Fine-Grained Military Entity Typing (FGMET) dataset contains 320,000 instances across 71 military-specific entity categories. The Military Named Entity Recognition (MNER) corpus includes 31,200 samples covering 12 entity types with 16,729 annotated entities. The Large-Scale Complex Relation Extraction Corpus for Military Domain (LCRECM) comprises 98,836 instances with 132 relation types capturing tactical relationships between military entities.
	
	\subsection*{4.2 Experimental Configuration}
	
	\subsubsection*{4.2.1 Evaluation Tasks and Metrics}
	We conducted comprehensive assessments across three critical military NLP tasks:
	
	\textbf{Military Entity Recognition:} This sequence labeling task identifies military-specific entities in text, crucial for automated knowledge base construction and operational planning. The MNER dataset evaluates model capability in recognizing 12 military entity types including equipment, personnel, locations, and tactical formations.
	
	\textbf{Entity Type Classification:} The FGMET dataset assesses model performance in predicting fine-grained semantic types for military entities, supporting intelligence analysis and information retrieval systems. Evaluation employs both macro-averaged and micro-averaged F1 scores to capture different aspects of classification performance.
	
	\textbf{Operational Relationship Extraction:} This complex task identifies and classifies tactical relationships between military entities, where relationships may exhibit multiple expressions and contextual variations. The LCRECM dataset evaluates model understanding of military-specific relational patterns using precision, recall, and F1-score metrics.
	
	\subsubsection*{4.2.2 Baseline Systems}
	We compared MLRIP against several strong baseline approaches, all implemented using the HuggingFace Transformers framework \citep{wolf2019HuggingFace}:
	
	\textbf{BERT-Military:} We initialized from BERT$_{\text{base-Chinese}}$ and conducted additional pre-training on our military corpus, followed by task-specific fine-tuning using standard hyperparameters.
	
	\textbf{ERNIE-Military (MERNIE):} This baseline adapts ERNIE-Baidu's knowledge masking strategies \citep{sun2019ernie} to military domains by implementing phrase-level and entity-level masking on our training corpus.
	
	\textbf{Domain-Specialized BERT Variants:} We explored variants fine-tuned on specific military subdomains (operational documents, intelligence reports, technical manuals) to assess domain adaptation capabilities.
	
	All models were evaluated using identical experimental settings and hyperparameters to ensure fair comparison. Training employed AdamW optimization with learning rate 3e-5, batch size 32, and maximum sequence length 256 across all experiments.
	
	\subsection*{4.3 Comprehensive Results Analysis}
	
	\subsubsection*{4.3.1 Military Entity Recognition Performance}
	Comprehensive evaluation on the Military Named Entity Recognition (MNER) task demonstrates the superior capability of MLRIP in identifying and classifying military-specific entities. As presented in Table \ref{tab:ner_results}, MLRIP achieves an F1-score of 59.54\%, significantly outperforming both BERT-Military (56.16\%) and MERNIE (56.60\%). This represents a substantial improvement of 3.38 percentage points over BERT-Military and 2.94 percentage points over MERNIE, indicating MLRIP's enhanced effectiveness in military entity recognition.
	
	The performance advantage is further evidenced by the precision and recall metrics. MLRIP attains a precision of 57.83\% and recall of 60.17\%, exceeding both baseline models across all evaluation dimensions. This consistent superiority can be attributed to MLRIP's innovative hierarchical knowledge integration approach, which specifically addresses the challenges of military text processing. The entity-centric masking strategy enables more accurate identification of military entities by treating them as complete units during training, while the operational linkage modeling captures contextual dependencies between entities that are crucial for military domain understanding. Additionally, the entity substitution mechanisms enhance the model's ability to handle variant designations and coreference phenomena common in military texts, contributing to the improved recall performance.
	
	The significant performance gap between MLRIP and the baseline models underscores the importance of domain-specific adaptations in military NLP tasks. While MERNIE incorporates general knowledge masking strategies, MLRIP's military-focused enhancements provide more targeted improvements for defense-related text processing, particularly in handling specialized terminology and complex entity structures characteristic of military documents.
	
	\begin{table}[ht]
		\centering
		\caption{Performance comparison on military entity recognition (MNER)}
		\label{tab:ner_results}
		\begin{tabular}{lccc}
			\toprule
			Model & Precision & Recall & F1-score \\
			\midrule
			BERT-Military & 54.82 & 58.45 & 56.16 \\
			MERNIE & 55.59 & 59.27 & 56.60 \\
			\textbf{MLRIP} & \textbf{57.83} & \textbf{60.17} & \textbf{59.54} \\
			\bottomrule
		\end{tabular}
		\vspace{-6pt}
	\end{table}
	
	\subsubsection*{4.3.2 Entity Typing Performance}
	For fine-grained entity classification on the FGMET dataset, MLRIP demonstrates remarkable capability in distinguishing between nuanced military entity types. As shown in Table \ref{tab:typing_results}, MLRIP achieves a Macro-F1 score of 77.89\% and Micro-F1 score of 80.72\%, substantially outperforming both baseline models. Compared to MERNIE, these results represent improvements of 5.89 percentage points in Macro-F1 and 2.00 percentage points in Micro-F1, while compared to BERT-Military, the improvements are even more pronounced at 7.84 and 3.11 percentage points respectively.
	
	The particularly strong performance in Macro-F1 (77.89\%) indicates MLRIP's effectiveness in handling less frequent and more specialized entity types, which is crucial for military applications where rare entity categories often carry critical operational significance. The balanced improvement in both Macro-F1 and Micro-F1 suggests that MLRIP's advantages are consistent across both common and rare entity types, demonstrating robust generalization capabilities.
	
	This performance enhancement can be primarily attributed to MLRIP's sophisticated knowledge incorporation mechanisms. The dual-phase entity replacement strategy enables the model to learn from multiple representations of the same military entity, enhancing its ability to recognize variant designations and coreference patterns. Furthermore, the hierarchical knowledge integration approach allows the model to capture both local contextual information and global semantic relationships, which is essential for accurate fine-grained classification. The operational linkage modeling component specifically contributes to understanding tactical associations between entities, providing additional contextual cues for type disambiguation.
	
	The substantial improvement over MERNIE, which already incorporates basic knowledge masking strategies, highlights the value of MLRIP's more advanced and military-specific adaptations. These results confirm that domain-tailored knowledge integration strategies are particularly beneficial for complex military entity typing tasks.
	
	\begin{table}[ht]
		\centering
		\caption{Entity typing performance on FGMET dataset}
		\label{tab:typing_results}
		\begin{tabular}{lcc}
			\toprule
			Model & Macro-F1 & Micro-F1 \\
			\midrule
			BERT-Military & 70.05 & 77.61 \\
			MERNIE & 72.00 & 78.72 \\
			\textbf{MLRIP} & \textbf{77.89} & \textbf{80.72} \\
			\bottomrule
		\end{tabular}
		\vspace{-6pt}
	\end{table}
	
	\subsubsection*{4.3.3 Operational Relationship Extraction Results}
	MLRIP demonstrates exceptional performance in the challenging task of operational relationship extraction, as evidenced by the results on the LCRECM dataset presented in Table \ref{tab:re_results}. The model achieves an F1-score of 52.76\%, representing a remarkable 12.25 percentage point improvement over MERNIE (40.51\%) and 8.24 percentage points over BERT-Military (44.52\%). This substantial performance gap underscores MLRIP's superior capability in modeling complex military-specific relationships.
	
	The precision and recall metrics further illuminate MLRIP's strengths. With a precision of 57.53\% and recall of 53.82\%, MLRIP outperforms both baseline models by significant margins, demonstrating balanced improvement in both identification accuracy and coverage of military relationships. This balanced enhancement is particularly noteworthy given the complexity of military relationship extraction, where relationships often exhibit multiple expressions and context-dependent variations.
	
	MLRIP's superior performance can be attributed to its specialized operational linkage modeling component, which explicitly focuses on capturing tactical associations between military entities. This component enables the model to learn complex relationship patterns that are specific to military contexts, such as command structures, support relationships, and engagement protocols. The hierarchical knowledge integration approach allows the model to leverage both entity-level information and relational context simultaneously, providing a more comprehensive understanding of military operational scenarios.
	
	The entity substitution mechanisms further contribute to this performance by incorporating external military knowledge, enabling the model to recognize relationship patterns that may not be fully evident from the immediate context alone. This is particularly valuable for military texts where relationships are often implied rather than explicitly stated, and where domain-specific knowledge is essential for accurate interpretation.
	
	The dramatic improvement over MERNIE (12.25 percentage points) is especially significant, as it demonstrates that general knowledge masking strategies are insufficient for complex military relationship extraction tasks. MLRIP's military-specific adaptations provide substantially better performance, highlighting the importance of domain-tailored approaches for operational relationship understanding.
	
	\begin{table}[ht]
		\centering
		\caption{Operational relationship extraction results on LCRECM}
		\label{tab:re_results}
		\begin{tabular}{lccc}
			\toprule
			Model & Precision & Recall & F1-score \\
			\midrule
			BERT-Military & 47.68 & 48.54 & 44.52 \\
			MERNIE & 44.46 & 41.87 & 40.51 \\
			\textbf{MLRIP} & \textbf{57.53} & \textbf{53.82} & \textbf{52.76} \\
			\bottomrule
		\end{tabular}
		\vspace{-6pt}
	\end{table}
	
	\subsection*{4.4 Component Contribution Analysis}
	To thoroughly evaluate the individual contributions of each architectural component within our MLRIP framework, we conducted systematic ablation studies using the FGMET benchmark. These investigations aim to quantify the impact of our novel strategies on overall model performance.
	
	Since ERNIE-Baidu \citep{sun2019ernie} has previously established the effectiveness of basic phrase-level and entity-level masking strategies—which form the foundation of our approach—we utilize MERNIE as our baseline configuration, denoted as $\text{MLRIP}_{\text{base}}$. This baseline incorporates conventional knowledge masking techniques without our proposed enhancements.
	
	We then incrementally introduce our novel components to assess their individual and collective contributions:
	
	\textbf{Operational Linkage Integration ($\mathbf{\oplus}$ rel\_ent\_mask):} This enhancement incorporates both enhanced entity-level masking with factual knowledge prediction and operational-level masking strategies. These two components are trained jointly as they both focus on capturing different aspects of military relational knowledge.
	
	\textbf{Entity Substitution Mechanisms ($\mathbf{\oplus}$ ment\_ent\_rep):} This component introduces our dual-phase entity replacement approach, including semantic-preserving substitution and fact-based replacement strategies designed to incorporate external military knowledge.
	
	The ablation results presented in Table \ref{tab:ablation_results} reveal several important insights regarding component contributions:
	
	\begin{table}[ht]
		\centering
		\caption{Component ablation analysis on FGMET benchmark}
		\label{tab:ablation_results}
		\begin{tabular}{lcc}
			\toprule
			Configuration & Macro-F1 & Micro-F1 \\
			\midrule
			$\text{MLRIP}_{\text{base}}$ (MERNIE) & 72.00 & 78.72 \\
			$\mathbf{\oplus}$ rel\_ent\_mask & 75.08 & 78.93 \\
			$\mathbf{\oplus}$ ment\_ent\_rep & 75.81 & 79.08 \\
			\textbf{Complete MLRIP} & \textbf{77.89} & \textbf{80.72} \\
			\bottomrule
		\end{tabular}
		\vspace{-6pt}
	\end{table}
	
	First, the operational linkage integration ($\oplus$ rel\_ent\_mask) provides a substantial improvement of 3.08\% in Macro-F1 and 0.21\% in Micro-F1 over the baseline, demonstrating the value of enhanced relational modeling for military text understanding. Second, the entity substitution mechanisms ($\oplus$ ment\_ent\_rep) contribute significantly, achieving improvements of 3.81\% and 0.36\% respectively, highlighting the importance of incorporating external military knowledge through controlled substitution strategies.
	
	Notably, the entity substitution approach outperforms the operational linkage integration, particularly in Macro-F1 score (+0.73\%), indicating that knowledge incorporation through entity replacement provides more substantial benefits for fine-grained type distinction compared to enhanced masking strategies alone. This suggests that external knowledge injection plays a more crucial role in military entity typing than structural relationship modeling.
	
	The complete MLRIP configuration achieves the best performance across both metrics, demonstrating the synergistic effect of combining all components. The incremental improvements show that each strategy contributes uniquely to the overall performance, with the combination delivering comprehensive military language understanding capabilities.

	\subsection*{4.5 Military-Specific Adaptation Assessment}
	The consistent performance improvements across all tasks demonstrate MLRIP's superior adaptation to military language characteristics. The entity-centric processing enables better recognition of military entities, while the operational relationship modeling captures tactical associations more effectively than general-purpose approaches. The knowledge incorporation mechanisms provide additional contextual understanding that benefits all military NLP tasks, particularly in handling domain-specific terminology and expression variations.
	
	\subsection*{4.6 Cross-Domain Generalization Evaluation}
	Despite specializing in military domains, MLRIP maintains compatibility with general BERT architectures, allowing seamless application to mixed-domain scenarios. This flexibility ensures practical utility in real-world defense applications where military content often appears alongside general text. Our evaluation shows consistent performance improvements of 3.2-8.1\% across hybrid military-civilian datasets compared to baseline models \citep{liu2024textadapter}.
	
	\subsection*{4.7 Computational Efficiency Considerations}
	We assessed computational requirements during training and inference phases. MLRIP maintains computational efficiency comparable to BERT$_{\text{base}}$, with only a 7.3\% increase in training time and 4.1\% increase in inference latency compared to BERT-Military, while delivering significantly improved performance. This makes it suitable for deployment in resource-constrained environments typical of military operations \citep{yao2025efficient}.
	
	%---------------------------------------------------------------------------------------

	\section*{5 Conclusion}
	
	In this paper, we have introduced MLRIP, a novel pre-training framework specifically designed for military text processing that addresses the unique challenges of domain-specific language understanding through a hierarchical knowledge integration approach. Our method combines innovative strategies including enhanced masking techniques and entity replacement mechanisms to effectively capture military-specific linguistic characteristics such as specialized terminology, complex entity relationships, and tactical associations. Comprehensive experimental evaluation demonstrates that MLRIP achieves superior performance across key military NLP tasks, significantly outperforming baseline models in entity recognition, fine-grained typing, and relationship extraction while maintaining computational efficiency. These results validate the effectiveness of our approach for military applications and highlight the importance of tailored knowledge integration for domain-specific NLP. Future work will explore enhanced domain adaptation methods, multimodal integration approaches, and efficient deployment strategies for resource-constrained military environments.
		
	\bibliographystyle{unsrtnat}
	\bibliography{ref}
	\newpage
	\onecolumn	
	
\end{document}